\documentclass{article}

\usepackage[dblblindworkshop,final, nonatbib]{neurips_2026}
\usepackage{amsmath}
\usepackage{graphicx}
\usepackage{placeins}

\workshoptitle{The 6th Workshop on Mathematical Reasoning and AI}

\usepackage[utf8]{inputenc} 
\usepackage[T1]{fontenc}    
\usepackage{hyperref}       
\usepackage{url}            
\usepackage{booktabs}       
\usepackage{amsfonts}       
\usepackage{nicefrac}       
\usepackage{microtype}      
\usepackage{xcolor}         
\usepackage[normalem]{ulem}
\usepackage{environ}
\usepackage{longtable}
\usepackage{array}
\usepackage{pgfplots}
\pgfplotsset{compat=1.18}
\usepackage{courier} 
\usepackage{etoolbox}
\usepackage[backend=biber,style=numeric-comp,sorting=none,giveninits=true,
  maxbibnames=99,doi=true,url=true,isbn=false]{biblatex}
\AtEveryBibitem{\clearname{editor}}

\title{Studying the Effects of Task Diversity Using Small Mathematical Worlds}

\author{%
  Emily Xia Meng \\
  University of Washington, Seattle \\
  \texttt{xmeng05@uw.edu}
  \And
  Xuanyu Yang \\
  University of Washington, Seattle \\
  \texttt{xy223@uw.edu}
  \And
  Junaid Hasan \\
  University of Washington, Seattle \\
  \texttt{junaid01@uw.edu}
  \AND
  Michael Ruofan Zeng \\
  University of Washington, Seattle \\
  \texttt{zengrf@uw.edu}
  \And
  Henry Kvinge \\
  University of Washington \\
  Pacific Northwest National Laboratory \\
  \texttt{henry.kvinge@pnnl.gov}
}

\begin{document}

\maketitle

\begin{abstract}
  Basic mathematical tasks like modular addition have been a valuable tool for illuminating aspects of neural network behavior that are challenging to see in noisy, unverifiable, large-scale, real-world settings. These studies have typically focused on a single task. However, modern language models are exposed to a vast number of tasks despite being trained with a next-token prediction objective. In this work, we introduce two mathematical task ecosystems, \texttt{IntWorld} and \texttt{PermWorld}, built from distinct but related tasks on integers and permutations. Both are designed to simulate task diversity within a cohesive mathematical `world' where tasks are distinct but related. Because the datasets are procedurally generated, we can precisely vary the number of tasks, allowing us to begin to disentangle the role of task variety from other factors in training. After describing the datasets, we provide some initial results obtained from these datasets which support the idea that increasing task diversity increases the richness of a model's internal representations.
\end{abstract}

\section{Introduction}
\label{sec:introduction}

From basic arithmetic assignments to scheduling problems to writing letters of recommendation, language models see a vast number of tasks during pretraining. Disentangling how this impacts model training is thus important for understanding the language models we have now as well as paths to designing better models across modalities in the future. 

Simple synthetic tasks have historically been a fruitful place to explore training dynamics and representation structure. These settings have the benefit that (i) we can control the contents of the training data exactly, (ii) we already know of algorithmic approaches to solving the problems, and (iii) we can verify the correctness of model outputs. As an example of the power of these frameworks, we note that the phenomenon of grokking was discovered by researchers studying how neural networks learn tasks such as modular arithmetic~\cite{power2022}. Modular addition and multiplication have also provided a setting in which a learned algorithm was successfully decoded \cite{nanda2023}, \cite{chen2026}. These single-task analyses leave interactions between tasks unexplored, making it challenging to study the impact of task diversity in these settings.

In this work, we describe progress toward addressing this limitation through the development of two mathematical task ecosystems that are designed to simulate multi-task learning. One is based on integers and the other is based on permutations. Both integers and permutations are fundamental mathematical objects with a diverse array of tasks and properties associated with them. Most importantly, each of these tasks illuminates one facet of these mathematical objects. This allows us to design a diverse collection of interrelated tasks. Just as experience writing a cover letter may impact one's ability to write a letter of recommendation, the task of multiplying numbers is also related to the task of determining divisibility properties. Finally, since these datasets are procedurally generated, we can exactly control the number of tasks in a dataset. 

We focus on three questions: what properties are represented, which tasks support transfer learning, and how performance changes on longer inputs.

Our contributions include the two task ecosystems, experiments varying the number of training tasks, and comparisons that test the effects of related tasks and training exposure.

\section{Related work}
\label{sec:related_work}

The role of task diversity has been of interest to several subfields of machine learning. Multitask learning studies how shared representations can benefit related tasks~\cite{caruana1997}, while task-grouping experiments show that the choice of tasks affects whether joint training helps~\cite{standley2020}. In a controlled regression setting, \textcite{raventos2023} study how the number of pretraining tasks affects in-context learning. Our transfer experiments instead fine-tune model parameters. Procedurally generated mathematics benchmarks also cover multiple tasks~\cite{saxton2019}; here we vary the training task set to study what models learn about a common class of objects.

Our representation analysis draws on linear probing~\cite{hewitt2019, alain_bengio_2016} and centered kernel alignment (CKA)~\cite{kornblith2019}. We use probes to predict mathematical properties from hidden states and CKA to compare representations across models. Representation convergence has been studied across broad model families~\cite{huh2024} and, more recently, across disjoint tasks in a geographic setting~\cite{park2026}. For permutations, \textcite{scullen2026} study how input notation affects learned representations, while PermuFormer uses multi-task, multi-encoding pretraining~\cite{kvingepermuformer}. We keep one-line notation fixed and use exact relations between properties to study task composition and representation alignment.

\section{Task Ecosystems from Small Mathematical Worlds}
\label{sect:task-ecosystems-from-small-mathematical-worlds}

In this section, we summarize our two task ecosystems \texttt{IntWorld} and \texttt{PermWorld}, the former based on integers and the latter based on permutations. Because these two families of objects are fundamental in multiple fields of mathematics, physics, and computer science and have been studied comprehensively, it is possible to assemble a large number of distinct but related tasks. Further, unlike some other synthetic settings where tasks are arbitrary, the tasks we use genuinely define core aspects of these objects. The tasks of addition or multiplication of integers, their divisibility properties, or the notion of a greatest common divisor or least common multiple are core facets of how we understand these objects. In a sense, these tasks define a core part of the \emph{small mathematical worlds} defined by integers and permutations (here we use informal terminology introduced in \cite{kvinge2026can}).

We provide examples of some of these below and provide the full list in Appendices~\ref{app:intworld} and~\ref{app:permworld}. We plan to extend these collections in future work. Code and experiment records are available in the repository:\\ {\vspace{-0.2in}\begin{center}\href{https://anonymous.4open.science/r/small-mathematical-worlds-65D6/}{https://anonymous.4open.science/r/small-mathematical-worlds-65D6}.\end{center}}

\paragraph{\texttt{IntWorld}: Integer Tasks} 
We choose integers because of their fundamental role in counting, ordering, and discrete mathematics. Integer tasks have clear labels, and their input scales can be easily modified through the number of digits, making them ideal for studying length generalization. Our tasks include arithmetic such as addition and multiplication, list sorting, and digit-based operations such as digit sum and reversal. The complete task list is provided in Appendix ~\ref{app:int_pretraining-tasks} and ~\ref{app:int_held-out-tasks}. We use sixteen integer tasks for pretraining and reserve four tasks, including sorting, for transfer.

\paragraph{\texttt{PermWorld}: Permutation Tasks} Permutations describe orderings of a finite set and admit a rich collection of  \emph{permutation statistics}, perfect as model-training tasks. For example, $[3,1,4,2]$ has two descents, no fixed points, and a longest increasing subsequence of length two. We use 32 scalar properties of permutations of length $2\leq n\leq30$. Each replicate divides the properties into two disjoint pools of sixteen; a model trained on a prefix of one pool is evaluated on the other. Appendices~\ref{app:perm_pretraining-tasks} and ~\ref{app:perm_held-out-tasks} contain a list of all tasks.

For \texttt{IntWorld}, we use sixteen designated pretraining tasks and four related held-out transfer tasks. For \texttt{PermWorld}, each replicate partitions the 32 properties into two disjoint pools of sixteen, with each pool serving as the held-out set for models trained on the other. Within each pretraining pool, we use fixed nested task sets containing 1, 2, 4, 8, and 16 tasks, denoted $T_1$, $T_2$, $T_4$, $T_8$, and $T_{16}$.

Finally, we note that when varying the number of tasks there is a direct trade-off between exposure to specific tasks and the total training budget. When training on $T_4$ and $T_8$, one can either keep the number of training steps the same or keep the number of examples seen from each task the same, but not both. We have opted to keep the number of training steps the same across all $T_k$ in the main experiments, since differences in total training budget seemed a more concerning confounder than differences in exposure to individual tasks.

\section{Experiments}
\label{sect:experiments}

In this section, we describe some initial experiments with \texttt{IntWorld} and \texttt{PermWorld}. In our experiments we train small MLPs and transformers. The integer MLPs have one causal token-and-channel mixing block, width 256, and hidden width 1,024. We use Transformers with four blocks, width 256, eight heads, and feed-forward width 1,024. For \texttt{IntWorld}
 experiments, we use base 100 representations of integers, because this balances legibility with token volume. For \texttt{PermWorld}, we keep one-line notation fixed throughout; effects of varying permutation encodings are studied in \cite{scullen2026}.

 In our experiments we measure performance for both tasks seen during training and new tasks. By varying the number of tasks in the training set we simulate higher task diversity and can measure how this changes different properties of the model. 
 
The properties we focus on are (i) performance on seen and unseen tasks, (ii) whether the model has learned transferable features, and (iii) the linear decodeability of held-out mathematical properties. (i) is measured by accuracy. (ii) is measured by finetuning the models to new tasks with a small amount of data, and to measure (iii) we train linear probes for held-out tasks. CKA methods and additional results are described in Appendix~\ref{app:geometry}.

Our training routine uses 20,000 update steps with effective batch size 64. We report means and sample standard deviations over three model seeds for integers and three task-split/model-seed replicates for permutations. The complete setup is described in Appendix~\ref{app:methods}.


\paragraph{Linear probing improves with task count for Transformers.}

We freeze each base model and train linear ridge probes to predict targets from tasks excluded from pretraining. For \texttt{PermWorld}, we extract the final-layer hidden state at the input-boundary token, immediately before the property token. This representation does not include the property token or the answer tokens. We train the probes using labeled examples of the held-out properties while keeping the base models fixed. We center and scale the target values within each input-length group using probe-training statistics and evaluate the probes on independent test examples using \(R^2\).

\begin{figure}[!htbp]
    \centering
    \includegraphics[width=0.9\textwidth]{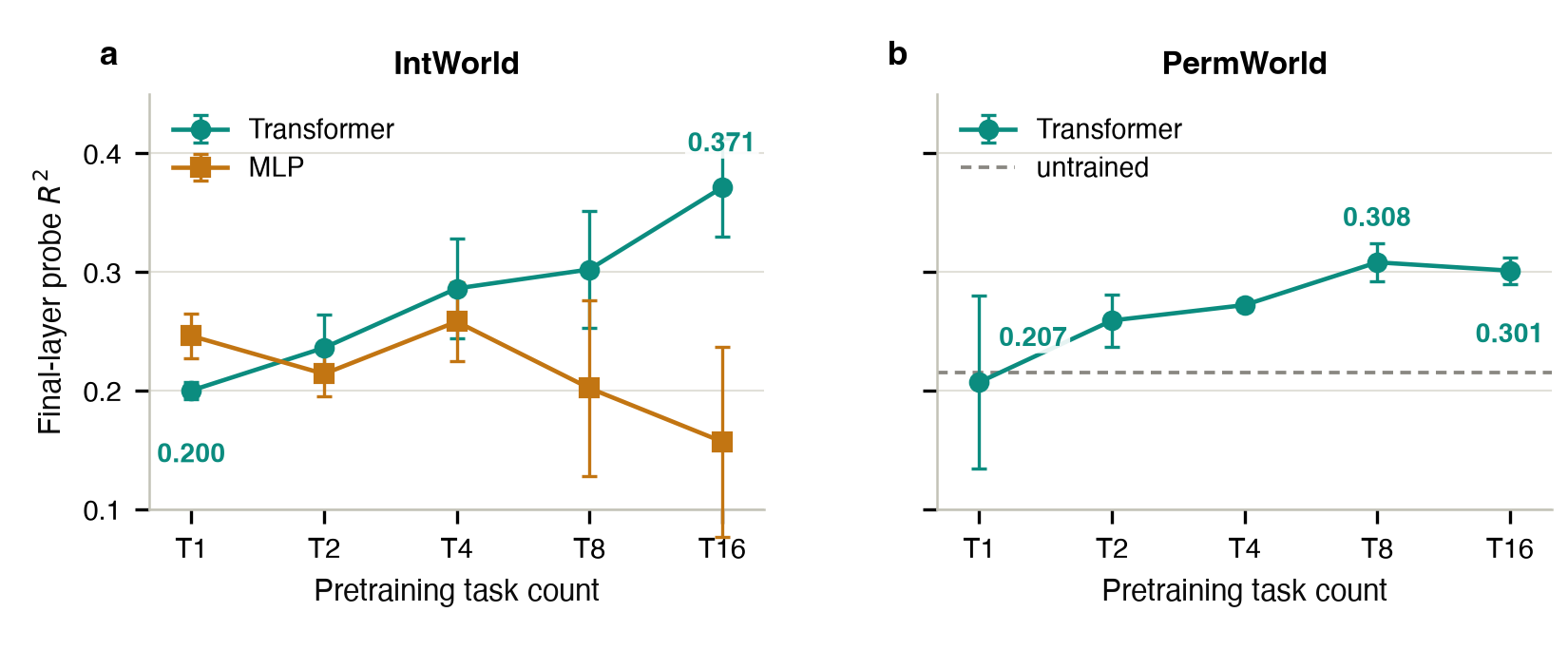}
    \caption{Linear probe performance generally improves with pretraining task diversity for Transformers}
    \label{fig:probe}
\end{figure}

For integer tasks, the $R^2$ score of the transformer final-layer probe increases steadily from $0.200$ at T1 to $0.371$ at T16 (Figure~\ref{fig:probe}a) suggesting that held-out properties are more linearly decodable in these broader training mixtures. The increase occurs for probes trained at other blocks as well (Appendix ~\ref{app:int_probing_layer}). The MLP does not show the same trend. For permutations, the probes predict properties excluded from pretraining: mean $R^2$ increases from $0.207$ at T1 to $0.308$ at T8, then decreases slightly to $0.301$ at T16 (Figure~\ref{fig:probe}b).


In the case of the transformer then at least, our results support (at a very small scale) the idea that task diversity is a path toward a model with richer internal representations. For \texttt{PermWorld}, the probe baseline from random initialization is $R^2=0.215$ (Appendix~\ref{app:perm-probes}). It is unclear why we do not see a similar trend for the MLP. It may be that the implicit token-based structure of the transformer makes it easier to build representations that are compatible across tasks. 

\paragraph{Transfer depends on task relevance.}

Multitask pretraining works best in few-shot transfer when pretraining included a directly related task. 

\begin{figure}[!htbp]
    \centering
    \includegraphics[width=0.9\textwidth]{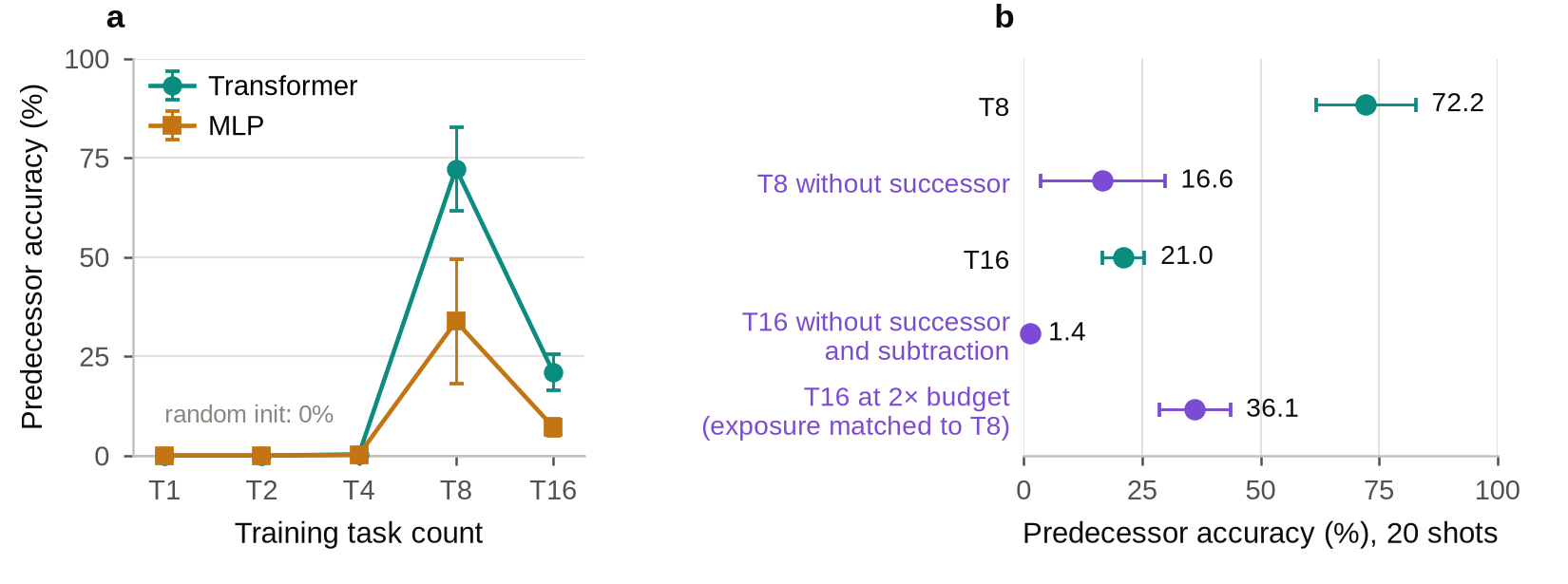}
    \caption{Few-shot predecessor transfer results}
    \label{fig:transfer}
\end{figure}

For \texttt{IntWorld}, we fine-tune on twenty examples of predecessor, a task absent from pretraining. The T8 Transformer achieves $72.24\pm10.51\%$ exact accuracy, compared with $0\%$ from random initialization and $0.27\pm0.27\%$ after T4 pretraining. Successor is included in T8 but not T4. Removing it from T8 reduces accuracy to $16.62\pm13.16\%$ (Figure~\ref{fig:transfer}), supporting the importance of this related task. Removing successor also changes the mixture size and per-task exposure. T16 performs worse than T8. Doubling its updates raises accuracy from $21.02\%$ to $36.12\%$, still below T8. The MLP model shows a similar but weaker trend in predecessor transfer. Transfer is more limited for the other holdouts: T8 twenty-shot accuracy is $0.85\%$ for least common multiple, $8.20\%$ for modular addition, and zero for sorting (Appendix~\ref{app:int_held-out-tasks}). 

For \texttt{PermWorld}, twenty-shot exact accuracy increased from
$16.64\pm2.60\%$ at $T_1$ to $33.59\pm3.61\%$ at $T_{16}$, while the gain over
zero-shot grew from $3.48\pm2.78$ to $17.80\pm5.76$ percentage points.
However, $T_{16}$ did not outperform random initialization, and because we
varied task count rather than the presence of a specific related property,
this indicates improved adaptability rather than a causal relevance effect.

\FloatBarrier

\FloatBarrier
\paragraph{Permutation few-shot transfer is sensitive to different learning rates}
For the \texttt{PermWorld}, our original few-shot experiment used different learning rates for the pretrained and randomly initialized models, which makes it difficult to separate the confounding effect of pretraining from the effect of the optimization setting. Therefore, to understand the results better, we repeated the experiment on validation data using the same learning rate for both groups. At T16, the pretrained model outperformed the randomly initialized model by $11.71\pm 5.24$ percentage points when both used a learning rate of $10^{-5}$, but by only $2.96 \pm 1.67$ percentage points when both used a learning rate of $3 \times 10^{-4}$ (Appendix Figure~\ref{fig:permutation-adaptation}). Hence, pretraining can improve twenty-shot performance on new, held-out permutation tasks, but this advantage is highly sensitive to the learning rate. Future experiments will test whether similar learning-rate sensitivity occurs in \texttt{IntWorld}.

\paragraph{From task-count scaling to combinatorial task geometry}
For the \texttt{PermWorld},
we use linear centered kernel alignment (CKA) ~\cite{kornblith2019}
to compare the models' internal representations. Let
\(X\in\mathbb{R}^{N\times d_X}\) and
\(Y\in\mathbb{R}^{N\times d_Y}\) be representation matrices whose
rows correspond to matched probe examples. We center each column using
\[
\widetilde{X}=HX,\qquad
\widetilde{Y}=HY,\qquad
H=I_N-\frac{1}{N}\mathbf{1}\mathbf{1}^{\top},
\]
and compute
\[
\operatorname{CKA}(X,Y)
=
\frac{\left\|\widetilde{X}^{\top}\widetilde{Y}\right\|_F^2}
{\left\|\widetilde{X}^{\top}\widetilde{X}\right\|_F
 \left\|\widetilde{Y}^{\top}\widetilde{Y}\right\|_F},
\]
where \(N=4096\) and \(\|\cdot\|_F\) denotes the Frobenius norm.
We test whether pairs of models trained on directly related properties
have higher CKA than pairs trained on properties with no specified
direct mathematical relation. For example, descents and recoils form a
related pair, whereas descents and fixed points do not belong to our prespecified pairs.

To test this, we first compare models trained on individual properties (Figure~\ref{fig:specialist-pairs}).
These properties form eight prespecified pairs with mathematical
relations, which can be found in Table~\ref{tab:relations}. The mean CKA is \(0.1375\) for the related pairs and \(0.0903\)
for the other pairs, with a one-sided task-label permutation test giving
\(p=0.015\). This suggests models trained using mathematically related properties tend to have more similar internal representations than the models trained using properties without a prespecified direct relation.

We then examine whether the higher CKA observed for related property pairs reflects the specific mathematical relationship between them (Figure~\ref{fig:symmetry}). 
For each related pair, we set three input conditions for the second model: no transformation, the correct inverse or complement transformation, and the alternative transformation. For example, we compare descents on a permutation with recoils on its inverse. The correct transformation produces higher CKA than both controls for all eight relations. Both relation-level two-sided exact sign tests give \(p=0.0078\). Together with the previous experiment, these results suggest that the representational similarity between related properties is sensitive to their specific mathematical relationship.

In the last experiment, we train each model on exactly four properties (Figure~\ref{fig:related-bundles}).
We vary the number of directly related property pairs between two
disjoint four-property training sets, using \(r\in\{0,1,2,4\}\) related
pairs. The average CKA is \(0.2935\), \(0.2752\),
\(0.2662\), and \(0.3648\), respectively. Only one of the 12 paired
curves increases monotonically. These results do not show a consistent
increase in CKA as the number of directly related property pairs
increases.
On unseen permutation tasks, $T_{16}$ validation zero-shot accuracy is $16.72\%$, below the $32.83\%$ majority-answer baseline (Appendix~\ref{app:perm-methods}).


\section{Conclusion}
\label{sec:conclusion}

Our initial experiments show how small mathematical worlds can be used to study task diversity. Transformer probe scores generally improve as the training task set grows, while successful transfer depends on the tasks learned together. Because the probe targets are excluded from pretraining, these results suggest that broader task exposure can make information about unseen mathematical properties more accessible in the learned representation. Behavior transfer is more conditional since it depends on which tasks are learned together and the benefit of pretraining also varies with the learning rate used during fine-tuning. Task count still changes per-task exposure, task composition, and how well the tasks are learned; our matched-exposure control covers only T16 predecessor transfer. In future work, we will vary task composition and training exposure independently and investigate how related tasks share computations.

\clearpage
\section*{References}
\printbibliography[heading=none]
\clearpage
\appendix
\section{Model Architecture and Training Setup}
\label{app:methods}

We use a common training setup across \texttt{IntWorld} and \texttt{PermWorld} to make comparisons between the two task ecosystems as direct as possible. For each domain, we train models on the nested task sets $T_1$, $T_2$, $T_4$, $T_8$, and $T_{16}$. Unless otherwise stated, each model is trained for 20,000 optimizer updates with AdamW and an effective batch size of 64. We use a learning rate of $3\times10^{-4}$, weight decay of $0.01$, 1,000 warmup steps, dropout of $0.1$, and clip the gradient norm at 1. Each task dataset contains 500,000 procedurally generated examples, partitioned into training, validation, and test sets using a $98/1/1$ split.

Our Transformer architecture consists of four pre-layer-normalized blocks with hidden width 256, eight attention heads, and GELU feed-forward layers of width 1,024. For \texttt{IntWorld}, we additionally train MLP models with one causal token-and-channel mixing block, width 256, and hidden width 1,024. Domain-specific tokenization and context lengths are described in the corresponding sections below.

We train three replicates for each task-count condition. For the integer experiments, the model seeds are 17, 42, and 314159. For the permutation experiments, each model seed is paired with a different partition of the 32 properties into disjoint pretraining and evaluation pools. Except for these task partitions and domain-specific input representations, the principal optimization settings and task-count conditions are shared across the two mathematical worlds.

\section{\texttt{IntWorld} Tasks}
\label{app:intworld}

\subsection{Experimental setup}
We trained models on nested subsets of the pretraining task collection in Appendix~\ref{app:int_pretraining-tasks}, using the task-count settings $T_1,\ldots,T_{16}$ introduced in Section~\ref{sect:experiments}, with both the Transformer and MLP architectures. The four tasks in Appendix~\ref{app:int_held-out-tasks} are excluded from pretraining and are selected to be closely related to specific pretraining tasks, enabling the relatedness ablations described below.

For $T_1$--$T_8$, training examples are drawn from an eight-task pool, yielding 500,000 examples per task and 4,000,000 examples in total. For $T_{16}$, we use an expanded sixteen-task pool with 8,000,000 examples. All sequences are capped at a 1,024-token context (Appendix~\ref{app:int_tokenization}).

The main experiment comprises $5$ task-count settings $\times$ $2$ architectures $\times$ $3$ seeds, for a total of 30 base models. We additionally train 9 control models to isolate task diversity from two confounds in the transfer experiments. These include $T_8$ and $T_{16}$ variants with the task(s) most closely related to the corresponding held-out predecessor task removed (successor, and successor plus subtraction, respectively), as well as a matched-exposure $T_{16}$ variant trained for 40,000 updates. The latter doubles the training budget so that its per-task exposure matches that of $T_8$.

\subsection{Integer representation and tokenization}
\label{app:int_tokenization}
We represent integers using base 100. 
For example, $7$ is \texttt{07}, while $247$ has base-100 digits \texttt{02 47}.
Integers greater than 99 are enclosed by \texttt{<NUM\_START>} and \texttt{<NUM\_END>}; integers from 0 to 99 use one number token. The most significant base-100 digit comes first. 
Each sequence contains a beginning-of-sequence token, a \texttt{<SIZE>} field, the input numbers, a task token, an equals sign, the answer, and an end-of-sequence token. Binary inputs delimit the second argument with \texttt{<ARG\_START>} and \texttt{<ARG\_END>}; tasks with an infix operator also include that operator.  
For example, $247+7=254$ is encoded as follows:

\begin{quote}\footnotesize\ttfamily
<BOS> <SIZE> 03 <NUM\_START> 02 47 <NUM\_END>\\
+ <ARG\_START> 07 <ARG\_END> <ADDITION> =\\
<NUM\_START> 02 54 <NUM\_END> <EOS>
\end{quote}
Training operands ordinarily have one to five decimal digits. Six-digit operands still fit in three base-100 number tokens, while seven-to-ten-digit operands require four or five. We evaluate these two ranges separately. Factorials are restricted to $0\leq n\leq99$ and are excluded from length extrapolation. For the digit-count task, the answer is already available in the \texttt{<SIZE>} field.

\subsection{Pretraining tasks}
\label{app:int_pretraining-tasks}
In the section below, \(n\), \(x\), and \(y\) denote non-negative integers, and \(m\) denotes a positive integer.

\textbf{Successor:} Given \(n\), return \(n+1\) \\
\textbf{Addition:} Given \(x\) and \(y\), return \(x+y\) \\
\textbf{Multiplication:} Given \(x\) and \(y\), return \(xy\) \\
\textbf{Modulo:} Given \(x\) and \(m\), return the remainder \(x\bmod m\) \\
\textbf{Greater-than comparison:} Given \(x\) and \(y\), return \(1\) if \(x>y\), and \(0\) otherwise \\
\textbf{Decimal digit sum:} Given \(n\), return the sum of its decimal digits. For example, the digit sum of \(247\) is \(2+4+7=13\), regardless of its base-100 tokenization \\
\textbf{Greatest common divisor:} Given positive integers \(x\) and \(y\), return \(\gcd(x,y)\) \\
\textbf{Integer-list sum:} Given a nonempty list \((x_1,\ldots,x_k)\), return \(\sum_{i=1}^{k}x_i\) \\
\textbf{Subtraction:} Given \(x\) and \(y\), such that \(x\geq y\), return \(x-y\) \\
\textbf{Integer division:} Given \(x\) and \(m\), return the quotient \(\left\lfloor \frac{x}{m}\right\rfloor\) \\
\textbf{Number of decimal digits:} Given \(n\), return the number of digits in its ordinary base-10 representation, with \(0\) treated as a one-digit integer \\
\textbf{Reverse decimal digits:} Given \(n\), reverse its ordinary base-10 digits and interpret the result as an integer. Leading zeros introduced by reversal are removed; for example, \(120\) returns \(21\) \\
\textbf{Decimal-digit occurrence count:} Given \(n\) and a digit \(d\in\{0,\ldots,9\}\), return the number of occurrences of \(d\) in the decimal representation of \(n\) \\
\textbf{Even/odd classification:} Given \(n\), return \(0\) if \(n\) is even and \(1\) if \(n\) is odd \\
\textbf{Divisibility classification:} Given \(x\) and \(m\), return \(1\) if \(m\mid x\), and \(0\) otherwise \\
\textbf{Factorial:} Given \(n\) with \(0\leq n \leq 99\), return \(n!=\prod_{i=1}^{n}i\), where \(0!=1\)

\subsection{Held-out transfer tasks}
\label{app:int_held-out-tasks}

In the section below, \(n\) denotes a positive integer, \(x\) and \(y\) denote non-negative integers, and \(m\) denotes a positive integer. These tasks are excluded from pretraining and used only for few-shot transfer evaluation. \\
\textbf{Predecessor:} Given \(n\), return \(n-1\). This task is closely related to successor. \\
\textbf{Least common multiple:} Given positive integers \(x\) and \(y\), return \(\operatorname{lcm}(x,y)\). This task is related to greatest common divisor and multiplication. \\
\textbf{Modular addition:} Given \(x\), \(y\), and \(m\), return \((x+y)\bmod m\). This task combines addition and modulo. \\
\textbf{Ascending sort:} Given a nonempty list \((x_1,\ldots,x_k)\), return the same elements in nondecreasing order. This task is related to comparison and list processing.

We fine-tune all model parameters on a fixed support set for 500 steps at learning rate $10^{-4}$ and select the checkpoint with the lowest validation loss. Pretrained and untrained models use the same support-set construction and selection rule. We evaluate twenty, fifty, and one hundred support examples per task.

Predecessor provides the strongest transfer result. With twenty examples, the T8 Transformer reaches $72.24\pm10.51\%$ exact accuracy, compared with zero from random initialization. Removing successor from T8 reduces accuracy to $16.62\pm13.16\%$. Doubling T16's pretraining updates raises its accuracy from $21.02\%$ to $36.12\%$. Figure~\ref{fig:transfer} shows these task-removal and training-budget controls. Since removing a task also changes the mixture size, both task identity and per-task exposure change in that comparison.

\begin{figure}[htbp]
\centering\includegraphics[width=\linewidth]{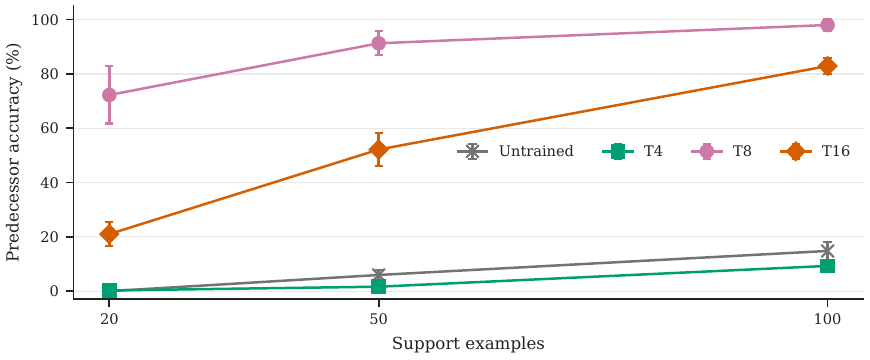}
\caption{Predecessor transfer by support size. Error bars show standard deviation across three seeds.}
\label{fig:integer-adaptation}
\end{figure}

With one hundred examples, T8 reaches $98.01\pm2.23\%$, compared with $82.91\pm3.08\%$ for T16 and $14.81\pm3.17\%$ from random initialization (Figure~\ref{fig:integer-adaptation}). The advantage is smaller on the other holdouts: T8 twenty-shot accuracy is $0.85\%$ for least common multiple, $8.20\%$ for modular addition, and zero for sorting. Sorting remains below $1\%$ even with one hundred examples.

\FloatBarrier
\subsection{Generalization by task and input length}
\label{app:integer-results}

\begin{figure}[htbp]
\centering\includegraphics[width=\linewidth]{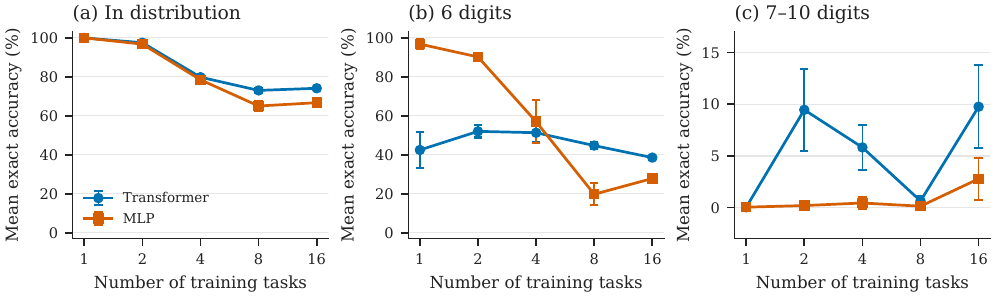}
\caption{Integer accuracy by input length. Means and SDs across three seeds.}
\label{fig:lengths}
\end{figure}

The task-count comparison changes both the training mixture and the number of presentations per task. Figure~\ref{fig:lengths} shows accuracy in distribution, at six digits, and at seven to ten digits. Neither architecture improves monotonically on the longest inputs. At T16, mean accuracy in that regime is $9.77\%$ for the Transformer and $2.78\%$ for the MLP.

\begin{figure}[htbp]
\centering\includegraphics[width=\linewidth]{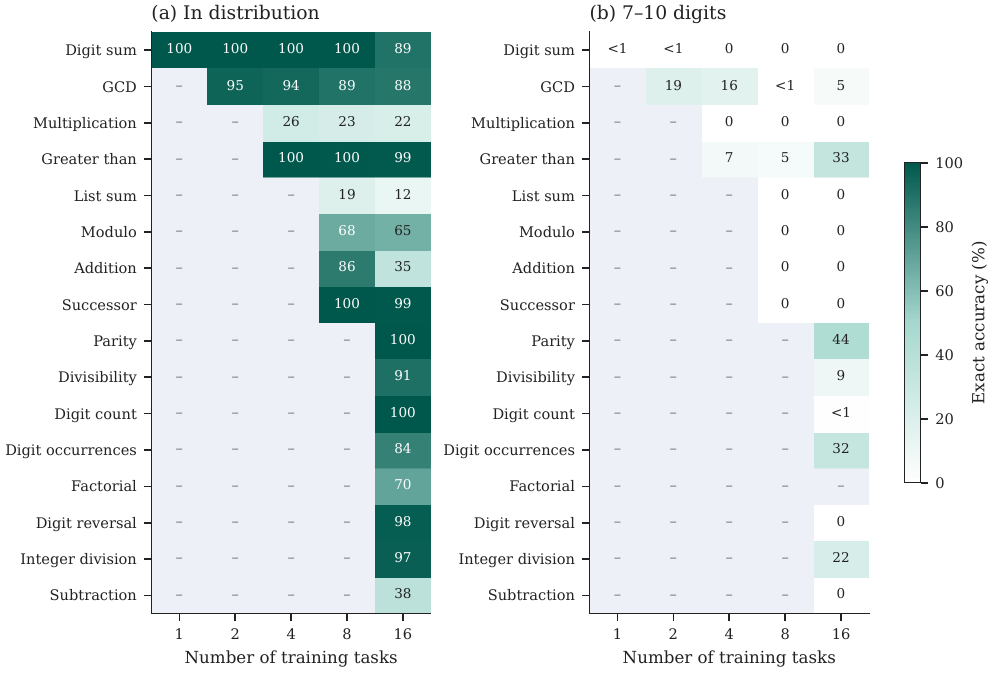}
\caption{Integer Transformer accuracy by task (\%). Cells average three seeds; $<1$ marks positive accuracy below $1\%$, and dashes mark unavailable evaluations.}
\label{fig:integer-tasks}
\end{figure}

Task-level results reveal differences hidden by the averages (Figure~\ref{fig:integer-tasks}). Digit sum, comparison, and successor have high in-distribution accuracy before T16, while multiplication and list sum perform substantially worse. Transformer addition accuracy falls from $86.04\pm8.11\%$ at T8 to $34.57\pm2.06\%$ at T16. On seven-to-ten-digit inputs, addition, multiplication, successor, and subtraction score zero in every mixture that includes them.

\subsection{Linear probing results by layer}
\label{app:int_probing_layer}
\begin{figure}[!htbp]
    \centering
    \includegraphics[width=\textwidth]{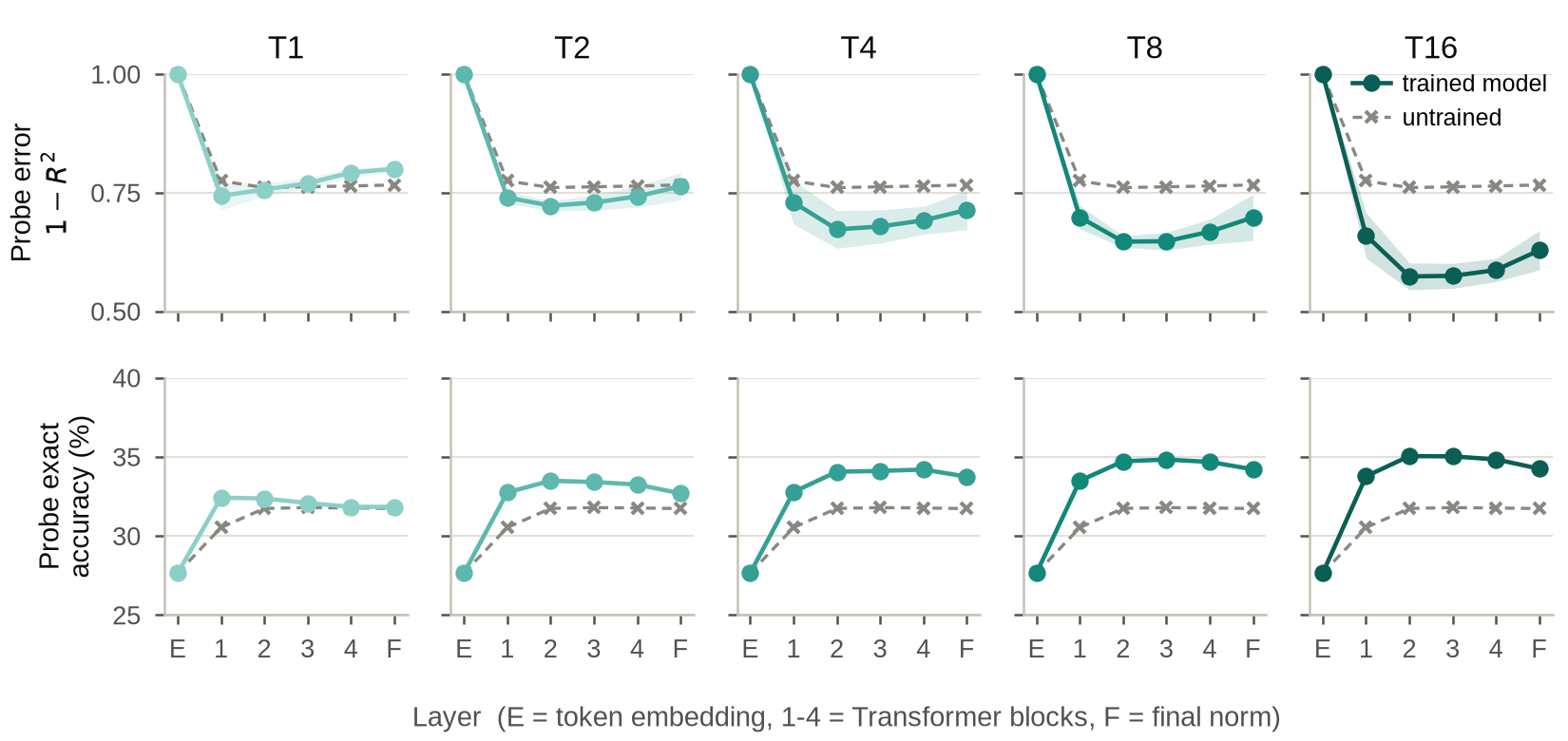}
    \caption{Linear probing by layer for Integer Transformer}
    \label{fig:int_probe_by_layer}
\end{figure}

We additionally evaluate linear probes at each stage of the Transformer. As shown in Figure ~\ref{fig:int_probe_by_layer}, probe performance is generally strongest at intermediate layers, particularly Blocks 2 and 3. After that, the performance decreases slightly toward the final normalized representation. The improvement with task diversity is also visible at intermediate layers, with probe performance generally improving as the number of pretraining tasks increases. 

\FloatBarrier

\FloatBarrier

\section{\texttt{PermWorld} Tasks}
\label{app:permworld}

\subsection{Pretraining tasks}
\label{app:perm_pretraining-tasks}

A permutation $\pi=[\pi_1,\ldots,\pi_n]$ is an ordered list that contains each integer from $1$ to $n$ exactly once. This is called its \emph{one-line notation}, which we use throughout. Table~\ref{tab:permutation-tasks} defines the 32 properties used in the scalar-task experiments. Subsequence lengths refer to strictly increasing or decreasing subsequences; runs must additionally be contiguous. Fixed points count as cycles of length one. All answers lie between zero and $n$, so one base-100 token suffices for the numerical answer.

\begingroup\small
\begin{longtable}{p{.28\linewidth}p{.65\linewidth}}
\caption{The 32 scalar permutation tasks. }\label{tab:permutation-tasks}\\
\toprule Property & Definition\\\midrule\endfirsthead
\toprule Property & Definition (continued)\\\midrule\endhead
\bottomrule\endfoot
\multicolumn{2}{l}{\textbf{Local order}}\\
Descents & $\#\{i<n:\pi_i>\pi_{i+1}\}$.\\
Recoils & Number of descents of $\pi^{-1}$.\\
Peaks & $\#\{1<i<n:\pi_{i-1}<\pi_i>\pi_{i+1}\}$.\\
Valleys & $\#\{1<i<n:\pi_{i-1}>\pi_i<\pi_{i+1}\}$.\\
Double ascents & $\#\{1<i<n:\pi_{i-1}<\pi_i<\pi_{i+1}\}$.\\
Double descents & $\#\{1<i<n:\pi_{i-1}>\pi_i>\pi_{i+1}\}$.\\
Successions & $\#\{i<n:\pi_{i+1}=\pi_i+1\}$.\\
Adjacencies & $\#\{i<n:|\pi_{i+1}-\pi_i|=1\}$.\\
\multicolumn{2}{l}{\textbf{Positions and records}}\\
Fixed points & $\#\{i:\pi_i=i\}$.\\
Anti-fixed points & $\#\{i:\pi_i=n+1-i\}$.\\
Excedances & $\#\{i:\pi_i>i\}$.\\
Deficiencies & $\#\{i:\pi_i<i\}$.\\
Left-to-right maxima & $\#\{i:\pi_i>\pi_j\text{ for every }j<i\}$.\\
Left-to-right minima & $\#\{i:\pi_i<\pi_j\text{ for every }j<i\}$.\\
Right-to-left maxima & $\#\{i:\pi_i>\pi_j\text{ for every }j>i\}$.\\
Right-to-left minima & $\#\{i:\pi_i<\pi_j\text{ for every }j>i\}$.\\
\multicolumn{2}{l}{\textbf{Cycles}}\\
Cycle count & Number of cycles in the disjoint-cycle decomposition.\\
Two-cycle count & Number of cycles of length two.\\
Three-cycle count & Number of cycles of length three.\\
Even-cycle count & Number of cycles of even length.\\
Odd-cycle count & Number of cycles of odd length.\\
Longest cycle & Maximum cycle length.\\
Shortest cycle & Minimum cycle length.\\
Nontrivial-cycle count & Number of cycles of length at least two.\\
\multicolumn{2}{l}{\textbf{Subsequences, runs, and global order}}\\
LIS length & Length of a longest increasing subsequence.\\
LDS length & Length of a longest decreasing subsequence.\\
Longest increasing run & Length of a longest contiguous increasing subsequence.\\
Longest decreasing run & Length of a longest contiguous decreasing subsequence.\\
Global descents & $\#\{i<n:\min_{j\leq i}\pi_j>\max_{j>i}\pi_j\}$.\\
Components & $1+\#\{i<n:\max_{j\leq i}\pi_j=i\}$; the number of direct-sum components.\\
Maximum displacement & $\max_i|\pi_i-i|$.\\
Displacement-one count & $\#\{i:|\pi_i-i|=1\}$.\\
\end{longtable}
\endgroup

Each of the three replicates uses a different partition into two pools. Prefixes of size four, eight, and sixteen are balanced across the four groups in Table~\ref{tab:permutation-tasks}. Each property changes pool in at least one replicate. No evaluation property occurs in its model's training pool. Related properties can occur across the two pools, as intended; for example, descents and recoils are related through permutation inversion. Independent record splits can still contain the same mathematical permutation, particularly at small $n$.

\subsection{Held-out transfer tasks}
\label{app:perm_held-out-tasks}

For the scalar-task experiments, the held-out tasks are the sixteen properties in the opposite pool. Thus the same property can be used for pretraining one model and for evaluating another. Within each replicate, the held-out set remains fixed as the training prefix grows from one to sixteen properties.

A separate twenty-task permutation experiment evaluates three holdouts: reduced-word translation, composition, and Lehmer-code translation. Reduced-word translation returns adjacent transpositions whose product is the input permutation; composition returns the one-line representation of a product of two permutations; and the Lehmer code records, at each position, the number of smaller entries to its right. These three tasks are excluded from pretraining in that experiment.

\subsection{Experiment description}\label{app:perm-methods}
For the permutation experiments, we partition 32 permutation properties into 2 disjoint 16-property pools and train them separately using four-layer Transformers on 1, 2, 4, 8, or 16 properties from each pool, denoted respectively as $T_1, T_2, T_4, T_8, $ and $T_{16}$. Permutation lengths range from 2 to 30, and each scalar answer occupies one number token. We use three random seeds, 17, 42, and 101, each paired with a different property partition. Five training task counts and two pools give thirty models. Each of the models is trained for 20,000 optimizer updates and evaluated on the 16 properties in the other pool. We keep the evaluation tasks fixed as the sixteen properties in the opposite pool for better comparison between models.

The corpus contains 500,000 examples per property, giving sixteen million records with a 98/1/1 train/validation/test split. The Transformer uses width 256, four pre-layer-normalized blocks, eight heads, learned positional embeddings, GELU feed-forward layers of width 1,024, and a context limit of 128 tokens. AdamW uses effective batch size 64, learning rate $3\times10^{-4}$, weight decay $0.01$, 1,000 warmup steps, dropout $0.1$, and gradient clipping at norm 1.

We report the original thirty-model experiment and a full retraining with the same configurations, task partitions, seeds, and corpus hashes. The per-task probe plots use the fully retrained models. Zero-shot accuracy averages 160 validation examples for each of the sixteen held-out properties. We first average properties and both pool directions within a replicate, then report means and sample SDs across the three joint task-split/model-seed replicates.

\subsection{Zero Shot Evaluation}
\begin{figure}[htbp]
\centering\includegraphics[width=\linewidth]{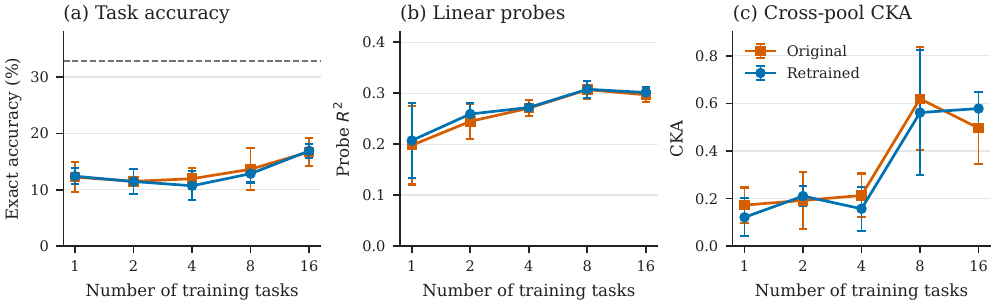}
\caption{Permutation results before and after retraining. The dashed line marks majority-answer accuracy. Error bars show SD across three random seeds.}
\label{fig:retraining}
\end{figure}

In the zero-shot evaluation, the average accuracy increases from $12.25 \pm 2.66\%$ with 1 training property to $16.72 \pm 2.50\%$ with 16 training properties. However, the accuracy does not increase consistently as the training properties increase and remains below the majority baseline of 32.83\% which predicts the most common answer for each task (Figure \ref{fig:retraining}). In the separate 20-task experiment, both Transformer and MLP models have 0\% accuracy on three unseen tasks: reduced-word translation, composition, and Lehmer-code translation. These results suggest that increasing the number of training properties is not enough for the models to reliably predict unseen tasks.

\FloatBarrier
\subsection{Linear probes}
\label{app:perm-probes}
To study the linear decodability of the model, we freeze the models and extract representations at every hidden layer for all thirty retrained permutation models. We then train linear ridge probes at each layer to see how well they could predict the properties that were not used to train the base models. The probes were fit and tuned on 4,096 validation permutations and evaluated using 4,096 independent test permutations. We normalized the target values within each permutation length using probe-training statistics to reduce the effect of length on the probe results.

As in \texttt{IntWorld}, we use a fixed 75/25 fitting/tuning split and select the ridge coefficient from $\{10^{-6},10^{-5},10^{-4},10^{-3},10^{-2},10^{-1},1,10\}$. For each target, we select the model trained on the opposite pool in each replicate. This gives three models per target and training task count, with the target held out at every layer. The individual-task plots report test $R^2$; the retraining comparison in Figure~\ref{fig:retraining} averages over the 32 targets and uses final-normalized representations.

The final-layer average $R^2$ on unseen properties is 0.2071, 0.2591, 0.2719, 0.3076, and 0.3013 for models trained on 1, 2, 4, 8, and 16 properties with all three seeds showing improvement from 1 to 8 training properties, and one run showing a decrease at 16 training properties. We also train probes using representations from randomly initialized Transformers, which give an average $R^2$ of 0.215 (Figure~\ref{fig:retraining}). The mean scores at T8 and T16 both exceed the untrained mean. The aggregate score improves from T1 to T8 and decreases slightly at T16; the individual-task plots show how this pattern varies across properties and layers.

We also examine probe performance across different layers at all five training task counts. For example, the mean probe $R^2$ for held-out descents at T16 rises from $0.164$ after block 1 to $0.258$ after block 4, then falls to $0.254$ after final normalization. Appendix~\ref{app:perm-task-probes} shows all 32 targets.

\FloatBarrier
\subsection{Few-shot transfer}
\label{app:adaptation}

Each base model is fine-tuned on four properties from its opposite pool, using twenty examples per property, all parameters trainable, batch size four, and 200 updates. These runs use the original base checkpoints. The 120 pretrained adaptations have 24 random-initialization controls. Final test evaluation uses 2,500 examples per property from a held-out source shard.

\begin{figure}[htbp]
\centering\includegraphics[width=\linewidth]{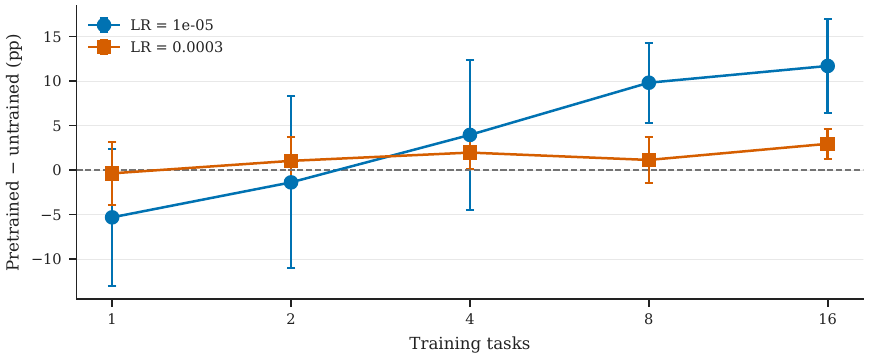}
\caption{Permutation adaptation at matched learning rates. Error bars show SD of paired accuracy differences.}
\label{fig:permutation-adaptation}
\end{figure}

In the original adaptation protocol, pretrained models use learning rate $10^{-5}$ and random controls use $3\times10^{-4}$. Test accuracy rises from $16.64\%$ at T1 to $33.59\%$ at T16, while random initialization reaches $34.28\%$. We therefore also compare pretrained and untrained models at matched learning rates on validation data (Figure~\ref{fig:permutation-adaptation}). At T16, the pretrained advantage is $11.71\pm5.24$ percentage points at $10^{-5}$ and $2.96\pm1.67$ at $3\times10^{-4}$. Error bars show the SD of paired accuracy differences across the three replicates.

\FloatBarrier
\subsection{Representation similarity}
\label{app:geometry}

We use linear centered kernel alignment (CKA; \cite{kornblith2019}) to compare the models' internal representations. We test whether models trained on related properties have higher CKA than models trained on other properties.

With column-centered feature matrices $X$ and $Y$ whose rows are paired input examples, linear CKA is
\[
\operatorname{CKA}(X,Y)=\frac{\|X^\top Y\|_F^2}{\|X^\top X\|_F\,\|Y^\top Y\|_F}.
\]
We use the final-normalized hidden state immediately before the task token. These comparisons relate our task-based experiments to work on representation convergence~\cite{huh2024,park2026} and permutation representation studies~\cite{scullen2026}.

\begin{figure}[htbp]
\centering\includegraphics[width=\linewidth]{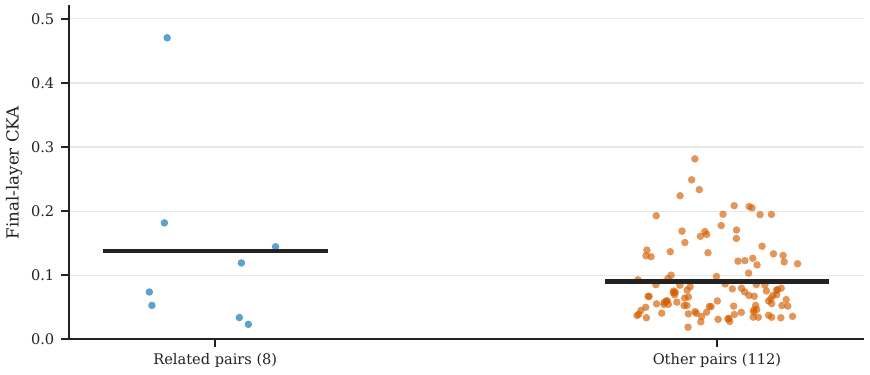}
\caption{Specialist CKA by task pair. }
\label{fig:specialist-pairs}
\end{figure}

To test this, we first compare models trained on individual properties. We use 16 properties and three random seeds, giving 48 single-property models. These properties form eight pairs with mathematical relations, such as descents and recoils. We extract final-layer representations before the property token using the same 4,096 validation permutations for every model. We then compare CKA for the eight related property pairs with the other 112 cross-property pairs. The average CKA is 0.1375 for the related pairs and 0.0903 for the other pairs, with a one-sided task-label permutation test giving $p=0.015$ (Figure~\ref{fig:specialist-pairs}).

We first average the six cross-seed comparisons for each distinct task pair. The permutation test holds this $16\times16$ task-similarity matrix fixed and shuffles the task labels 100,000 times, recomputing the difference between the eight designated related pairs and the other 112 pairs. Figure~\ref{fig:specialist-pairs} shows the task-pair means, including the variation within each group.

\begin{figure}[htbp]
\centering\includegraphics[width=\linewidth]{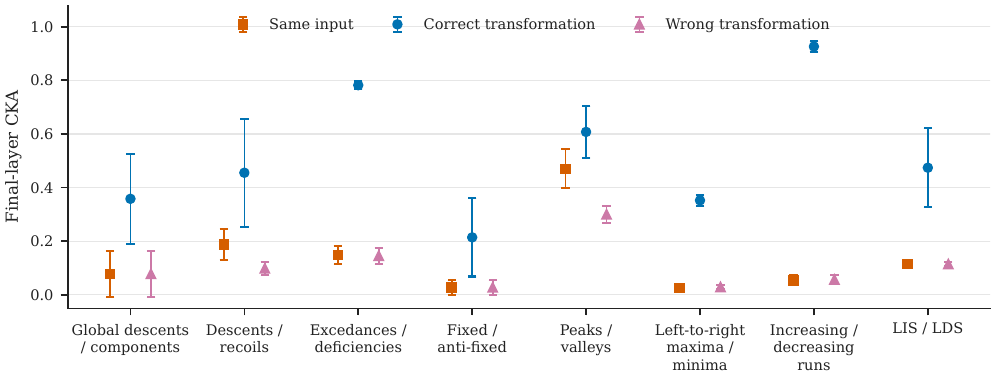}
\caption{Specialist CKA under the same input, the correct transformation, and a control transformation. }
\label{fig:symmetry}
\end{figure}

We also examine whether the alignment between related properties reflects the mathematical relationship between their inputs. We use the same single-property models and 4,096 validation permutations. For each related pair, we keep the input to the first model unchanged and compare three input conditions for the second model: no transformation, the correct inverse or complement transformation, and the other transformation as a control. For example, for a correct inverse pair, we give a permutation to the descents model and its inverse to the recoils model. The correct transformation gives higher CKA than both control conditions for all 8 property relations across the three seeds (Figure~\ref{fig:symmetry}). Both comparisons have a relation-level two-sided sign-test $p=0.0078$. These results suggest that the models' representations reflect mathematical relationships between properties.

For each sign test, we average over seeds within a mathematical relation and use the eight relation means as the units. The correct-transformation contrast is positive for all eight relations against either control, giving $p=2/2^8=0.0078125$. Table~\ref{tab:relations} lists the transformations. Each identity was checked exhaustively on permutations of lengths two through eight.

\begin{table}[htbp]
\centering\small
\caption{Relations used for the specialist comparison. Value complement is $c(\pi)_i=n+1-\pi_i$. The transformed right-hand property equals the left-hand property, except in the last row where it is larger by one.}
\label{tab:relations}
\begin{tabular}{ll}\toprule
Property pair & Transformation\\\midrule
Descents / recoils & Inverse\\
Peaks / valleys & Value complement\\
Fixed / anti-fixed points & Value complement\\
Excedances / deficiencies & Inverse\\
Longest increasing / decreasing subsequence & Value complement\\
Longest increasing / decreasing run & Value complement\\
Left-to-right maxima / minima & Value complement\\
Global descents / components & Value complement\\\bottomrule
\end{tabular}
\end{table}

\begin{figure}[htbp]
\centering\includegraphics[width=\linewidth]{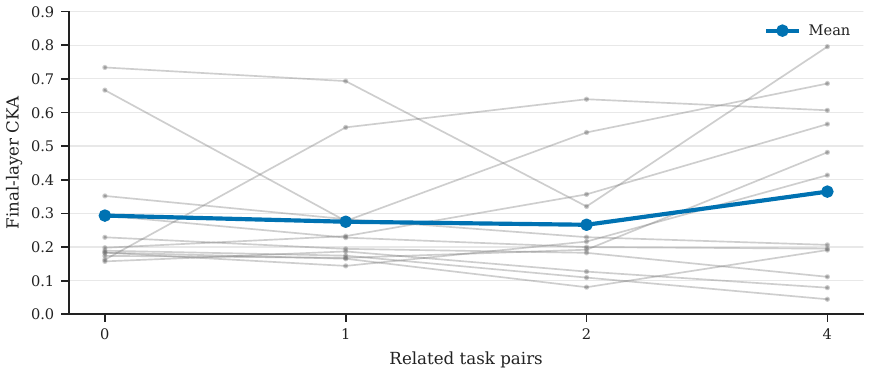}
\caption{CKA with four training tasks per model. Thin lines show the twelve layout--seed combinations; the thick line shows their mean.}
\label{fig:related-bundles}
\end{figure}

In a separate experiment, we keep 4 training properties for each model. We vary the number of directly related property pairs between the two disjoint training pools, using 0, 1, 2, or 4 related pairs. The average CKA is 0.2935, 0.2752, 0.2662, and 0.3648, respectively (Figure~\ref{fig:related-bundles}). These results do not show a consistent increase in CKA as the number of related pairs increases.

This experiment crosses four task-pool layouts with three model seeds, giving twelve curves. Only one curve increases monotonically across all four conditions. The mean paired difference between four and zero related pairs is $0.0712\pm0.2121$, with seven of twelve differences positive. The two-sided exact sign test gives $p=0.7744$. The input transformations therefore provide clearer evidence of alignment than the number of related training pairs alone.

\FloatBarrier
\subsection{All-layer probe plots}
\label{app:perm-task-probes}

The following two pages cover all 32 permutation properties, with sixteen targets per page. Each panel fixes one property. Curves show T1, T2, T4, T8, T16, and untrained models, with means and sample SDs across three joint task-split/model-seed replicates. Emb. denotes the embedding output and Norm. final normalization. All panels use the same $R^2$ scale.

\clearpage

\begin{figure}[htbp]
\centering\includegraphics[width=\linewidth]{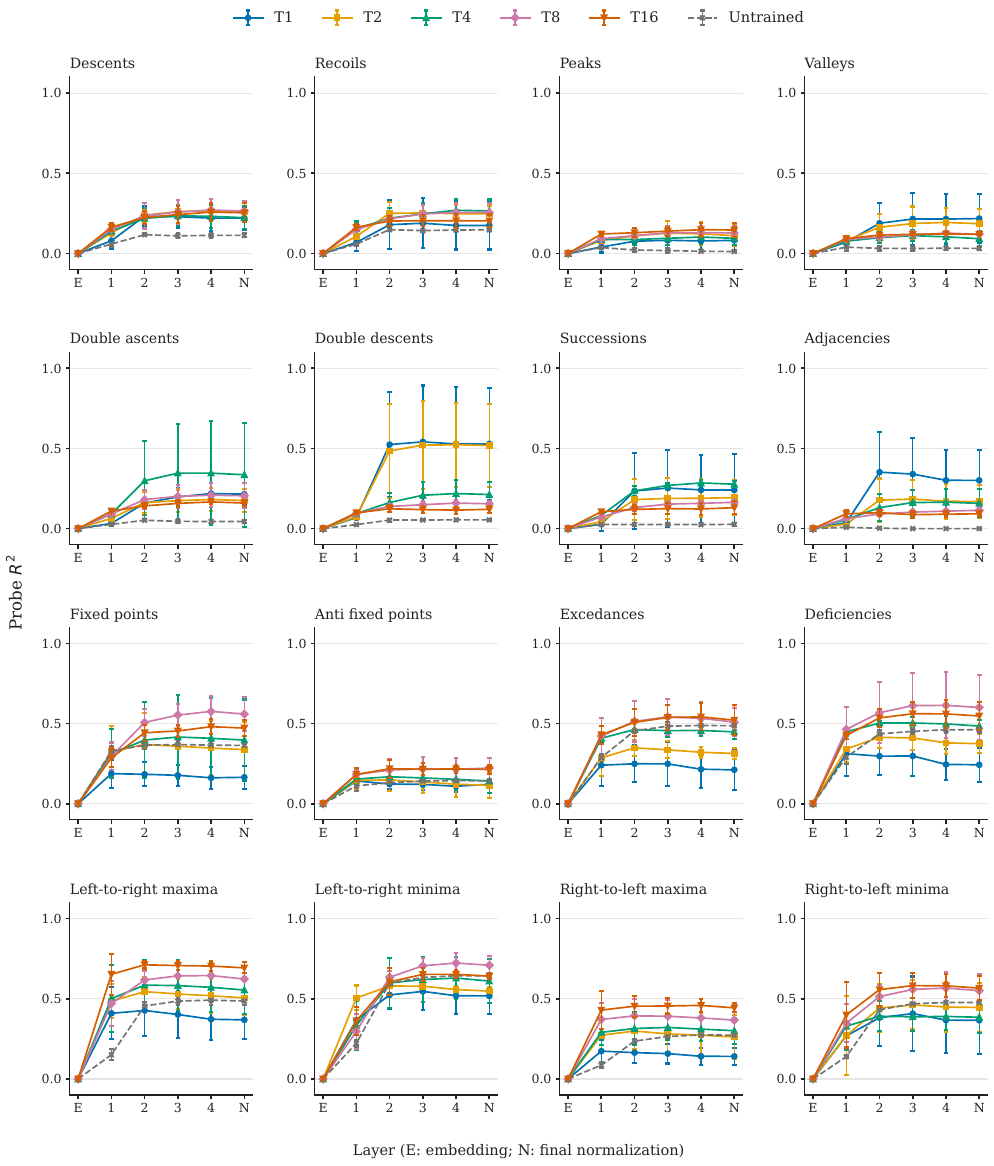}
\caption{Permutation probes: order and positions.}
\label{fig:permutation-sheet-01}
\end{figure}

\clearpage

\begin{figure}[htbp]
\centering\includegraphics[width=\linewidth]{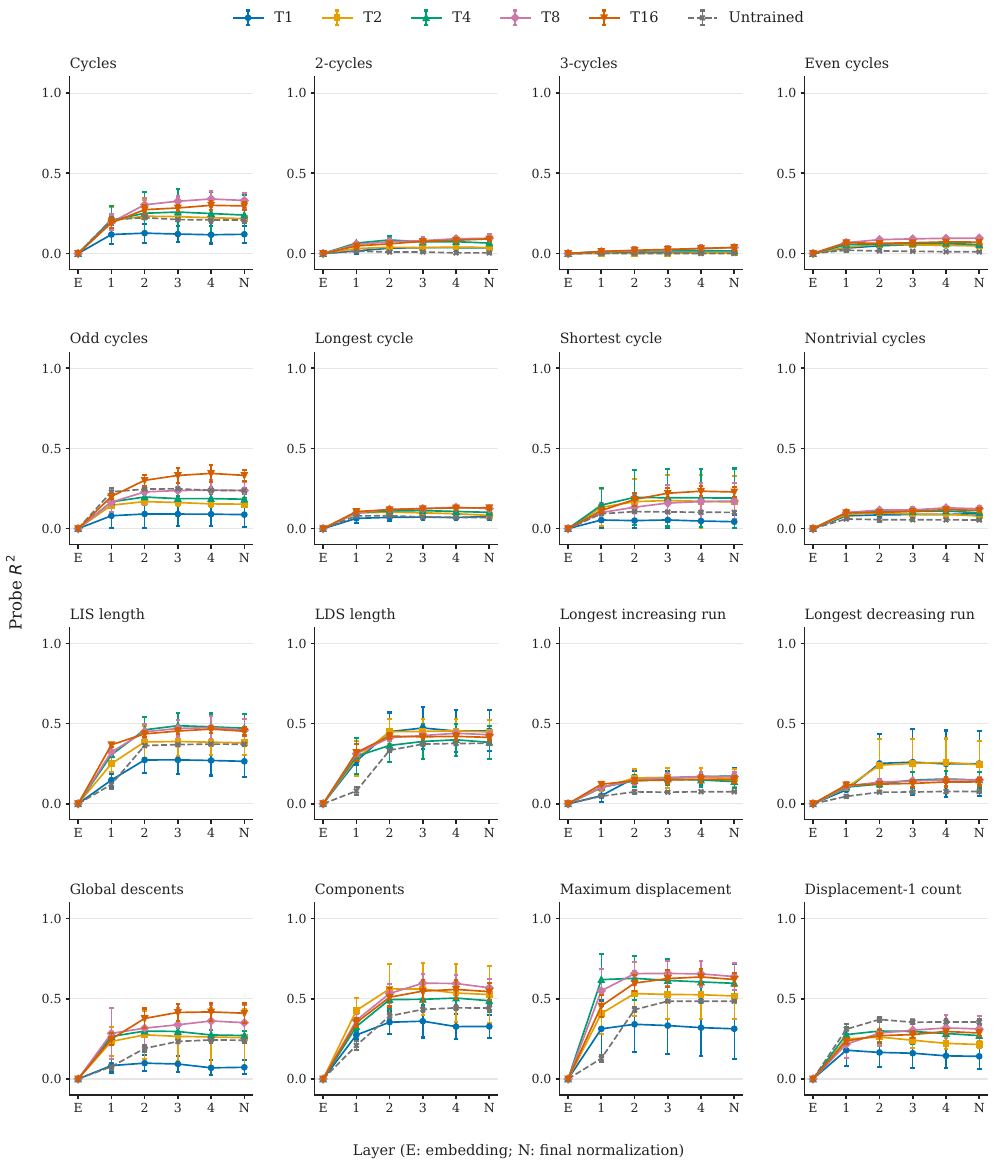}
\caption{Permutation probes: cycles and global structure.}
\label{fig:permutation-sheet-02}
\end{figure}

\end{document}